\documentclass[10pt,twocolumn,letterpaper]{article}

\usepackage[final]{cvpr}      
\definecolor{cvprblue}{rgb}{0.21,0.49,0.74}
\usepackage[pagebackref,breaklinks,colorlinks,allcolors=cvprblue]{hyperref}

\def\paperID{} 
\def\confName{CVPR}
\def\confYear{2026}

\title{Giraffe: A Mapping Architecture from Hidden Text Representations to Visual Embeddings for Efficient Graphic Design}

\author{Nejla Ghaboosi\\
Canva Research\\
{\tt\small nejla@canva.com}
}

\begin{document}
\maketitle
\begin{abstract}
Multimodal large language models (MLLMs) have made significant progress in understanding and interpreting multimedia content. However, their ability to generate media remains limited. Recent approaches have attempted to bridge this gap by translating the hidden representations of token sequences into the embedding space of visual models or directly into raw image data. However, these methods often represent each image using multiple specialised tokens which significantly increases the input length. This becomes a major limitation for tasks such as graphic design generation where the output typically involves a seamless blend of thousands of tokens across text, multiple images, and layout information. To address this challenge, a novel architecture is proposed that maps hidden token representations to the embedding space of visual models, such as CLIP ViT-L/14, using a single \texttt{[IMG]} token per image. The architecture employs two shallow MLP blocks, each with a separate compression module followed by a shared expansion module, trained with six distinct loss functions. One block aids the other during training and is omitted during inference, resulting in a lightweight solution. Strong performance is demonstrated in both image-to-design and text-to-design generation tasks.
\end{abstract}
    
\section{Introduction}
\label{sec:intro}

Multimodal large language models (MLLMs) have made remarkable advancements in tasks involving visual comprehension and understanding such as visual question answering (VQA), image captioning, and object grounding \cite{b1, b2, b3, b4, b5, b6}. However, these methods primarily focus on processing multimodal inputs but only generating textual outputs. This restricts their effectiveness in tasks that require generating visual content alongside text, such as graphic design generation. Graphic designs are composed of various multimodal elements including text, images, SVG shapes, colours, and spatial layout, all working together to convey meaning and visual appeal.

\begin{figure}[t]
  \centering
  \includegraphics[width=\columnwidth]{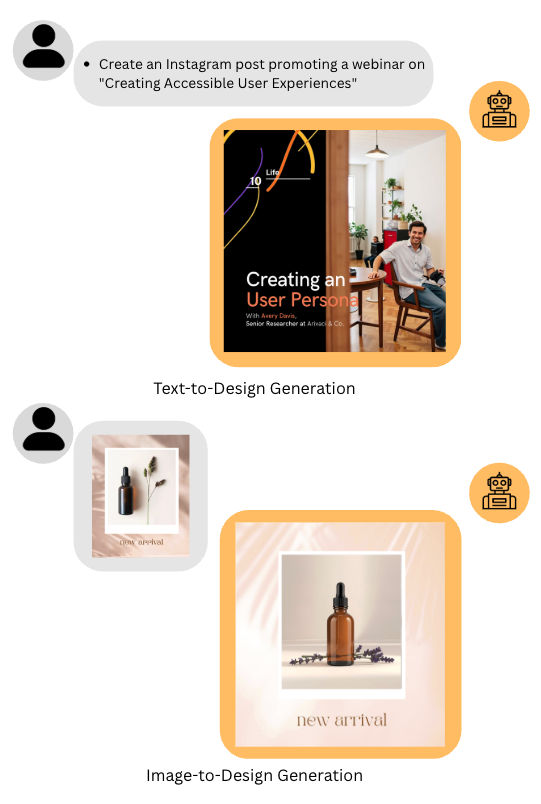}
  \caption{Example results from the proposed architecture. The model generates visually coherent and stylistically consistent graphic designs from either text or image inputs.}
   \label{fig:fig0}
\end{figure}

Recent approaches have explored the expansion of language models to accommodate the generation of multimodal outputs. In FROMAGe \cite{b7}, image retrieval capabilities are introduced by augmenting a frozen language model with two trainable linear layers. The first linear layer maps the hidden representation of a \texttt{[RET]} token whilst the second maps the visual embedding of the corresponding image. These two embeddings are aligned through contrastive learning \cite{b8}, allowing the model to associate textual queries with relevant visual content.  Nonetheless, this approach falls short when a design requires an image with a specific subject, style, or colour theme that is not available in the media library, or when user lacks access to a sufficiently large media collection.

GILL \cite{b9} extends the capabilities of FROMAGe by enabling media generation rather than just retrieval. This is achieved through a mapping network called GILLMapper - a transformer with both encoder and decoder components. GILLMapper translates the hidden representations of \textit{r} \texttt{[IMG]} tokens, generated by the language model, into the embedding space of Stable Diffusion’s \cite{b10} text encoder. The alignment between the two embedding spaces is learned using a mean squared error (MSE) loss. During inference, the images are generated by passing the mapped embeddings through the diffusion model. 

Similarly,  Emu \cite{b11} enables media generation by augmenting a language model with a module called Causal Transformer that transforms visual inputs into \textit{N} \texttt{[IMG]} visual tokens. The model is trained to autoregressively predict these visual tokens, using an MSE loss between the predicted embeddings and the ground-truth tokens generated by the Causal Transformer. During inference, the generated visual tokens are decoded into images using a Stable Diffusion model that has been fine-tuned to explicitly condition on generated embeddings. 

In Chameleon \cite{b12},  images are represented using 1024 discrete tokens derived from a codebook of size 8192, produced by a learned image tokenizer. These image tokens, along with text tokens, are used to train the model autoregressively. During inference, a learned de-tokenizer is used to reconstruct images from generated image tokens.

Across all of these methods, a shared characteristic is the use of multiple visual tokens to represent images. In fact, \cite{b9} further investigates the impact of token count and finds that reducing the number of tokens generally leads to performance degradation as it produces shorter and less expressive inputs for the mapping network.This issue becomes particularly significant in graphic design generation tasks, where a single design may contain multiple images and is typically represented as a long, interleaved sequence of text and image tokens to capture elements such as text, images, SVG shapes, layout, and visual attributes. Consequently, representing each image with multiple tokens can cause the overall token sequence to become extremely long, often approaching or exceeding the model’s maximum token capacity. Moreover, the limited attention span of transformer-based architectures makes it challenging to capture long-range dependencies across such extended sequences, thereby constraining the model’s ability to produce coherent and consistent designs.

Some approaches have explored generating images directly rather than relying solely on embeddings. For instance, Transfusion \cite{b13} leverages a VAE \cite{b14} to encode images into latent representations and then applies a diffusion process within this latent space. In this setup, text is modelled autoregressively using a transformer while image generation is guided by a diffusion-based objective.  DreamLLM \cite{b15}, on the other hand, enables direct image generation by using a frozen Stable Diffusion decoder that is guided by semantic queries produced by the language model. These queries are trained using Score Distillation Sampling (SDS) \cite{b16}, allowing the model to align textual prompts with image generation targets. Like the aforementioned methods, both these approaches still rely on multi-vector or multi-token representations per image, which may limit their effectiveness in design generation tasks.

\begin{figure*}[t]
  \centering
  \includegraphics[width=\textwidth]{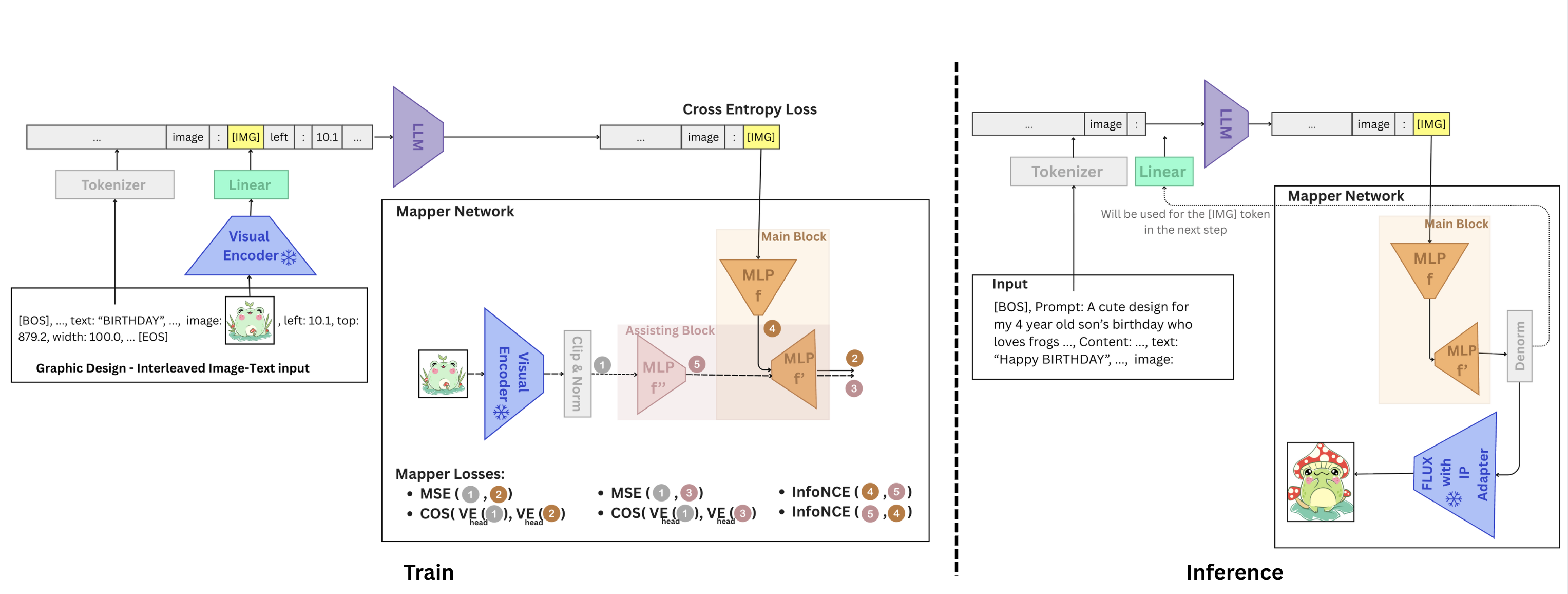}
  \caption{The proposed Giraffe architecture overview. It enables mapping from hidden token representations to the embedding space of visual models, such as CLIP ViT-L/14, using only a single \texttt{[IMG]} token. This is achieved using two shallow MLP blocks where each includes a seperate compression module followed by a shared expansion module, trained with six distinct loss functions (left). The assisting MLP block aids the main block in learning the mapping and is ignored during inference (right), leading to a more lightweight solution.}
   \label{fig:fig1}
\end{figure*}

This paper proposes a novel architecture that facilitates mapping from hidden token representations of a language model to the embedding space of a visual model, like CLIP ViT-L/14 \cite{b17}, using just a single \texttt{[IMG]} token per image. This enables the language model to create coherent graphic designs by integrating text and image tokens within long sequences. The proposed mapping architecture, termed Giraffe due to its L-shaped structure, is composed of two shallow MLP blocks. Each block features a unique compression module, followed by a shared expansion module that links the two. One block is mainly responsible to handle the mapping, while the other assists the main block during the training phase, playing an essential role in helping the primary block to reach its mapping goal. When it comes to inference, the assisting block is removed, leading to a more lightweight solution. Since the generated embeddings follow a known distribution, there is no need to fine-tune a diffusion model. An existing pretrained model such as FLUX \cite{b18} with CLIP VIT-L/14 IP Adapter \cite{b19} can be used directly to generate images. 

Extensive experiments highlight its strong performance in both text-to-design generation and image-to-design tasks. In the text-to-design generation task, it can be seen that the produced images align with the provided prompt while preserving the overall style, color palette, and semantics of the graphic design, leading to an aesthetically pleasing outcome. Likewise, in the image-to-design task, the generated graphic design closely mirrors the reference image. Collectively, the observed behavior indicates that the proposed architecture effectively captures all image-related information within a single \texttt{[IMG]} token. \cref{fig:fig0} shows an example for each task. It is important to note that while this paper focuses primarily on images, the proposed architecture is flexible and can be adapted for various media, including audio and video.

\section{Method}
This section describes the graphic design representation used by the model, the proposed Giraffe architecture, the training objectives, and the inference process.

\subsection{Graphic Design Structure}

A graphic design is composed of various elements, including text, images, and SVG shapes, which all work together to convey ideas and create visual harmony. Each element is defined by a set of properties that determine its structure and appearance. These include positional attributes such as top, left, width, height, and layering order, as well as visual characteristics like colour, transparency, and rotation. Certain attributes are also specific to particular element types. For example, text elements may include the textual content itself, font family, font size, and styling options such as bold or italic. By representing these element-specific properties in textual form, a graphic design can be represented as a long sequential structure of interleaved text and image components, where each $x_t$ and $x_v$ denotes a single text or image component, respectively. The inclusion of separate image components is necessary, as visual content cannot be fully captured through text alone.

\subsection{Giraffe Architecture}

Following prior work \cite{b1, b3, b20, b21, b22}, the visual embedding of each image $x_v$ is extracted and projected into the same dimensional space as the word embeddings, using the following transformations:
\begin{equation}
\begin{aligned}
z_v &= v_{\phi}(x_v)  \in  \mathbb{R}^d, \quad \\
u_v &= Wz_v  \in  \mathbb{R}^m
\end{aligned}
\end{equation}
, where $W$ is a learnable linear mapping, $v_{\phi}(.)$ is the pretrained and frozen visual encoder (ViT-L/14) prior to its projection head and $m$ represents the input embedding dimension of the language model.
Each $x_v$  in the input sequence is then replaced with just one special \texttt{[IMG]} token and its corresponding word embedding is substituted with $u_v$.

To train the model to understand and generate multimodal content, it is optimised with the standard next-token prediction objective. Specifically, the model is fine-tuned to predict each next token using the cross-entropy loss:
\begin{equation}
\begin{aligned}
\mathcal{L}_{\text{CE}} = -\frac{1}{M} \sum_{m=1}^{M} \sum_{t=1}^{T_m} \log P_{\psi,W} \left( w_t^{(m)} \mid w_{<t}^{(m)} \right)
\end{aligned}
\end{equation}
, where $M$ is the batch size, $T_m$ is the length of the $m$\text{th} input sequence, and $\psi$ denotes the language model parameters. Each token $w_t$ may be either a text token or the special \texttt{[IMG]} token.

Finally, to enable image generation, a mapping network is integrated with the language model to decode each \texttt{[IMG]} token into its corresponding normalised visual embedding $\tilde{z}_v$. $\tilde{z}_v$ is obtained by first clipping the original visual embedding $z_v$ to remove extreme values, followed by applying the min-max-normalision technique to rescale the values into the range [-1, 1]. This mapping is learned using a multilayer perceptron (MLP) block, which comprises of a compression module $f_{\theta_1}$ followed by an expansion module $ f'_{\theta_s}$ with SiLU activation function applied throughout both modules. However, the last layer in $f'_{\theta_s}$ uses Tanh as its activation function. The parameters of this block are optimised by minimising the mean square error (MSE) loss between its output and $\tilde{z}_v$:

\begin{equation}
\begin{aligned}
\mathcal{L}_{\text{b1-MSE}} &= \frac{1}{N} \sum_{i=1}^N \left( \hat{\tilde{z}}_{\text{v}_i} - \tilde{z}_{\text{v}_i} \right)^2, \\
\hat{\tilde{z}}_{\text{v}_i} &= f'_{\theta_s}(f_{\theta_1}(h_\psi(\text{[IMG]}_i)))
\end{aligned}
\end{equation}
, where $h_\psi([IMG]_i)$ denotes the representation of the $i\text{th} $ \texttt{[IMG]} token from the final hidden layer of the language model and $N$ represents the number of \texttt{[IMG]} tokens in the batch.

Additionally, a cosine similarity loss is used to promote directional alignment between the predicted embeddings and the ground truth:
\begin{equation}
\begin{aligned}
\mathcal{L}_{\text{b1-COS}} &=  \frac{1}{N} \sum_{i=1}^N \left( 1 - \frac{g_\varphi(\hat{\tilde{z}}_{\text{v}_i}) \cdot g_\varphi( \tilde{z}_{\text{v}_i})}{\|g_\varphi(\hat{\tilde{z}}_{\text{v}_i})\| \|g_\varphi( \tilde{z}_{\text{v}_i}) \|} \right)
\end{aligned}
\end{equation}
, where $g_\varphi(.)$ denotes the projection head of the pretrained and frozen ViT-L/14 visual encoder. $g_\varphi(.)$ is applied to both predicted and target embeddings to ensure compatibility in the representation space. 

While the proposed MLP block establishes the baseline mapping, it often struggles to learn the mapping effectively. To enhance this block's ability to fully capture the transformation, an assisting MLP block is introduced, forming the Giraffe architecture named for its L-shaped structure. The assisting block features its own compression module $f''_{\theta_2}$ with SiLU as activation function but shares the expansion module $f'_{\theta_s}$ with the main MLP block. Functionally, it operates as an autoencoder where the input is the normalised visual embedding $\tilde{z}_v$ and the output is its approximation $\bar{z}_{\text{v}_i}$. Similarly, it is trained with two losses:
\begin{equation}
\begin{aligned}
\mathcal{L}_{\text{b2-MSE}} &= \frac{1}{N} \sum_{i=1}^N \left( \bar{z}_{\text{v}_i} - \tilde{z}_{\text{v}_i} \right)^2, \\
\bar{z}_{\text{v}_i} &= f'_{\theta_s}(f''_{\theta_2}(\tilde{z}_{\text{v}_i}))
\end{aligned}
\end{equation}
and,
\begin{equation}
\begin{aligned}
\mathcal{L}_{\text{b2-COS}} &= \frac{1}{N} \sum_{i=1}^N \left(1 - \frac{g_\varphi(\bar{z}_{\text{v}_i}) \cdot g_\varphi( \tilde{z}_{\text{v}_i})}{\|g_\varphi(\bar{z}_{\text{v}_i})\| . \|g_\varphi( \tilde{z}_{\text{v}_i} \|} \right).
\end{aligned}
\end{equation}

To further assist $f_{\theta_1}$ in learning the mapping transformation, two infoNCE losses \cite{b8} are applied on the bottlenecks of the two MLP blocks:
\begin{equation}
\begin{aligned}
\mathcal{L}_{\text{NCE}_1}& = -\frac{1}{N} \sum_{i=1}^N \left( log\frac{e^{sim\left(f_{\theta_1}(\tilde{z}_{\text{v}_i}), f''_{\theta_2}(\tilde{z}_{\text{v}_i})\right)/\tau}}{\frac{1}{M} \sum_{j=1}^M e^{sim\left(f_{\theta_1}(\tilde{z}_{\text{v}_i}), f''_{\theta_2}(\tilde{z}_{\text{v}_j})\right)/\tau}} \right), \\
\mathcal{L}_{\text{NCE}_2} &= -\frac{1}{N} \sum_{i=1}^N \left( log\frac{e^{sim\left(f_{\theta_1}(\tilde{z}_{\text{v}_i}), f''_{\theta_2}(\tilde{z}_{\text{v}_i})\right)/\tau}}{\frac{1}{M} \sum_{j=1}^M e^{sim\left(f_{\theta_1}(\tilde{z}_{\text{v}_j}), f''_{\theta_2}(\tilde{z}_{\text{v}_i})\right)/\tau}} \right),
\end{aligned}
\end{equation}
where $\tau$ is a learnable temperature parameter. 

The total loss is calculated as:
\begin{equation}
\begin{aligned}
\mathcal{L}_{\text{total}} &= \lambda_1\mathcal{L}_{\text{b1-MSE}} + \lambda_2\mathcal{L}_{\text{b1-COS}}  \\
&+ \lambda_3\mathcal{L}_{\text{b2-MSE}} + \lambda_4\mathcal{L}_{\text{b2-COS}} \\
&+ \lambda_5(\mathcal{L}_{\text{NCE}_1}  + \mathcal{L}_{\text{NCE}_2}) +  \lambda_6\mathcal{L}_{\text{CE}},
\end{aligned}
\end{equation}
where $\lambda_1$, $\lambda_2$, $\lambda_3$, $\lambda_4$, $\lambda_5$ and  $\lambda_6$ are hyperparameters representing loss weights. The optimisation is carried out over parameters $W$, $\theta_1$, $\theta_2$, $\theta_s$, and $\psi$. An overview of the architecture is shown in \cref{fig:fig1}.

\subsection{Inference}

At inference time, $f''_{\theta_2}$ is removed, resulting in a lightweight mapping network. Each predicted \texttt{[IMG]} token is mapped into its normalised visual embedding as follows: 
\begin{equation}
\begin{aligned}
\hat{\tilde{z}}_{\text{v}} &= f'_{\theta_s}(f_{\theta_1}(h_\psi(\text{[IMG]}))),
\end{aligned}
\end{equation}
which is then denormalised to recover the predicted visual embedding $\hat{z_\text{v}}$. During the autoregressive process, $\hat{z_\text{v}}$ is used directly instead of recomputing $z_v$ via the visual encoder $v_{\phi}(.)$ for each generated \texttt{[IMG]} token, enabling faster token generation.

Finally, real images are produced by passing $\hat{z}_{\text{v}}$ to FLUX with CLIP VIT-L/14 IP Adapter. The generated images are then placed within the specified bounding boxes to create the final graphic design.

\section{Experiments}

To evaluate the proposed method, two experiments are carried out on text-to-design and image-to-design generation tasks. The model is trained on a proprietary dataset of 1,800,000 professionally created graphic designs. These graphic designs span a wide range of formats, including social media posts, banners, flyers, business cards, and logos. The dataset is split into training and validation subsets, with 90\% used for training and 10\% for validation. 

In the mapping network, $f_{\theta_1}(.)$ consists of layers with dimensions $1024-768-512$, while $f''_{\theta_2}(.)$ has dimensions $1024-768-512-512$, with the final layer in both serving as the bottleneck. $f'_{\theta_s}(.)$ is composed of layers with dimensions $768-1024-1024$.  The hyperparameters $\lambda_1$, $\lambda_2$, $\lambda_3$, $\lambda_4$, $\lambda_5$ and  $\lambda_6$ are set to  1, 1, 1, 1, 0.1, and 1, respectively.

\subsection{Text-to-design generation}

\begin{table}[t]
\caption{FID score and CLIP cosine similarity for text-to-design and image-to-design tasks, respectively.}
\centering
\label{tab:table1}
\begin{tabular}{|p{2cm}|c|c|}
\hline
\textbf{Model} & \multicolumn{1}{c|}{\textbf{Text-to-Design}} & \multicolumn{1}{c|}{\textbf{Image-to-Design}}\\
 & \textbf{(FID)} & \textbf{(Cosine)}\\
\hline
Proposed & 66.85 & $0.85 \pm 0.05$    \\ \hline
Baseline & 81.61 &  -  \\ \hline
\end{tabular}
\end{table}

\begin{figure*}[t]
  \centering
  \includegraphics[scale=0.57]{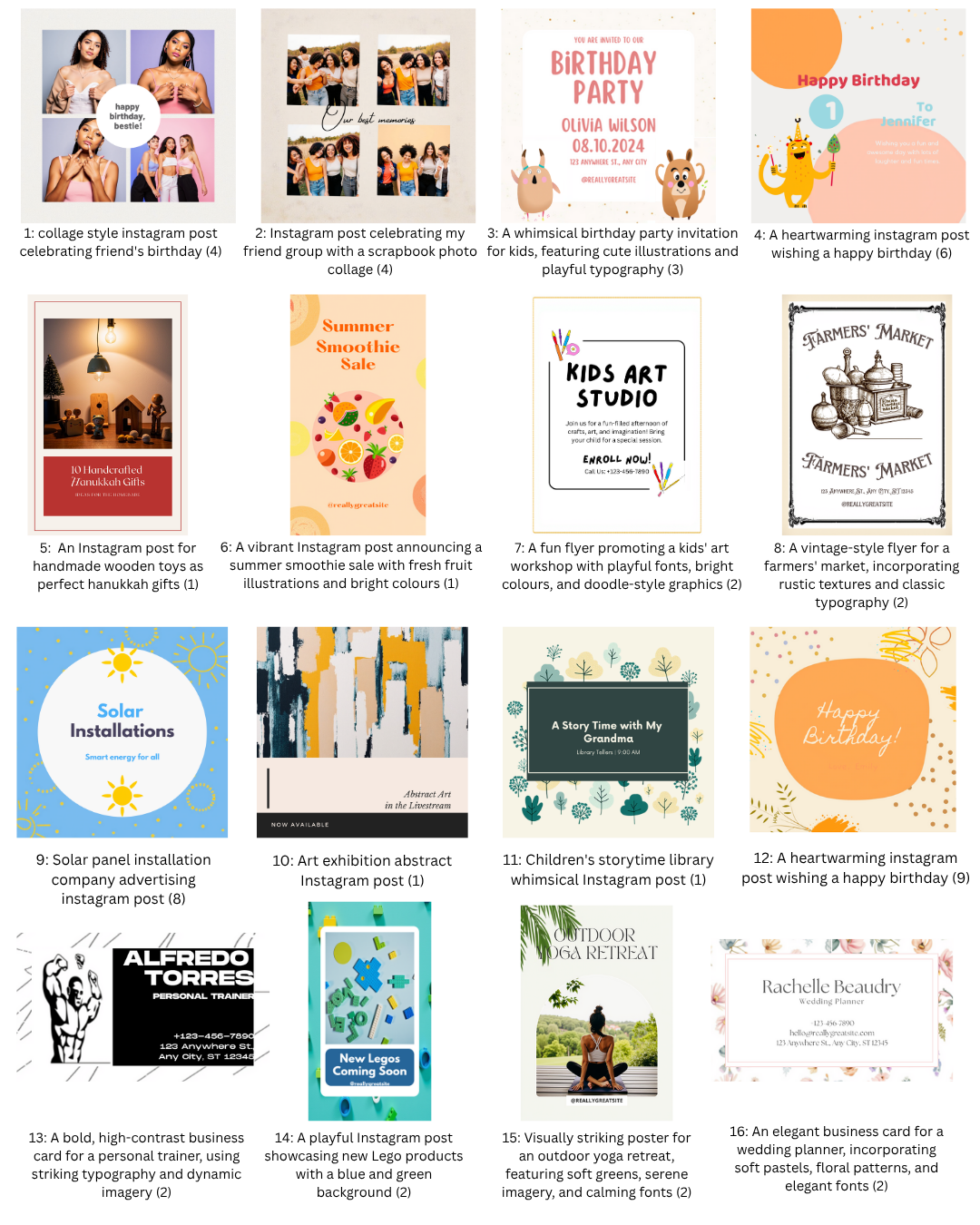}
  \caption{Qualitative results for the text-to-design task. The generated graphic designs closely follow the prompts while remaining visually coherent, stylistically consistent, and of high overall quality. The model often produces complex designs with multiple images that interact harmoniously. Numbers in parentheses indicate the number of images used in each design.}
  \label{fig:fig2}
\end{figure*}

\begin{figure*}[t]
  \centering
  \includegraphics[scale=0.567]{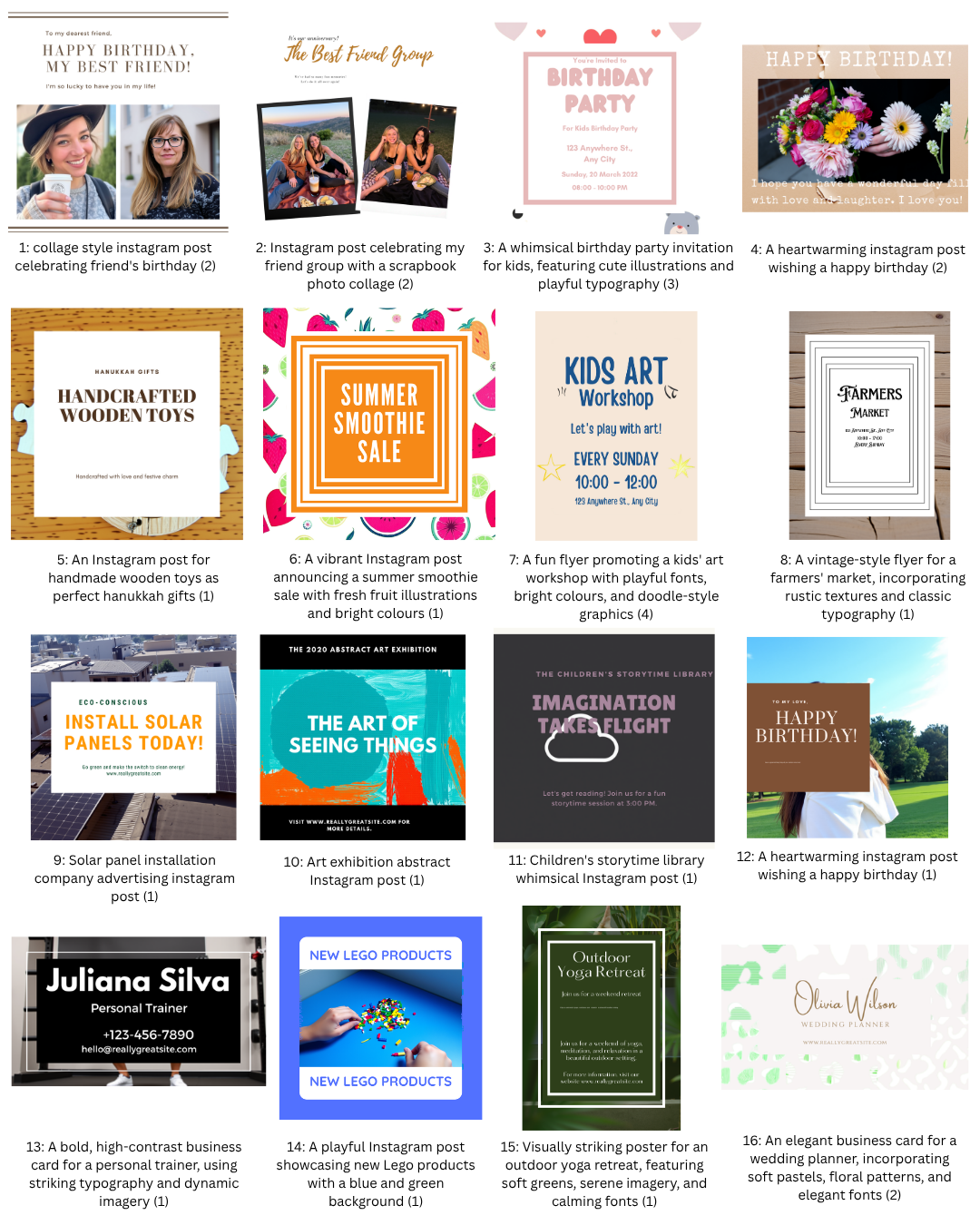}
  \caption{Qualitative results for the baseline model. Generated results have limited layout variation and less consistent visual style. Numbers in parentheses indicate the number of images used in each design.}
  \label{fig:fig3}
\end{figure*}

In this experiment, the Gemma3 4B model \cite{b23} is used as the language model. Training is carried out in mixed precision using the bfloat16 \cite{b24} format for 56,000 steps. Evaluation is conducted on a separate set of curated prompts, chosen to cover a wide range of cases. Each design sample is represented in JSON format, with floating-point bounding box coordinates rounded to the nearest integer. Furthermore, each design is accompanied by a textual description that captures its style, theme, and key details, for example, “A vibrant Instagram post announcing a summer smoothie sale with fresh fruit illustrations and bright colours”. For benchmarking, a baseline model is also trained without the mapping network, in which textual description of each image is generated instead and subsequently passed to FLUX for image generation.

To evaluate the performance of the model, a comprehensive qualitative analysis is performed on the generated results. \Cref{fig:fig2} and \cref{fig:fig3} show sample outputs generated by the proposed architecture and the baseline model, respectively, for the same set of prompts. The number of images in each design is indicated in parentheses. As shown in these figures, the proposed method generates designs that are more visually appealing, with greater layout diversity and stronger stylistic and thematic coherence. Many of the generated designs include multiple images that work harmoniously in terms of color theme and style, as seen in prompts 3, 4, 9, and 12. In contrast, the baseline model often produces designs with minimal layout variation, featuring a background image, a central shape, and overlaid text. When the baseline generates designs containing multiple images, the images usually lack color harmony and appear visually not very relevant to the overall design. For instance, in prompt 7, the images generated by the baseline model do not exhibit consistent color or style, whereas the corresponding design produced by the proposed method maintains a coherent theme, with images that complement each other and enhance the overall design coherence. These limitations in the baseline model likely arise from representing each image solely through textual descriptions. The resulting increase in token length, together with the limited attention span of transformer-based architectures, makes it difficult to capture long-range dependencies and constrains the model’s ability to generate more complex designs. In addition, representing images as text alone makes it challenging to accurately capture their style and color theme. In contrast, the \texttt{[IMG]} token represents each image with a single token, allowing the model to focus on finer-grained details within its attention span. Moreover, the \texttt{[IMG]} token preserves the style and thematic information of an image more faithfully, enabling the proposed model to generate outputs that are more complex, visually coherent, stylistically consistent, and better aligned with the intended design.

To further evaluate the performance of the proposed mapping network, the FID score \cite{b25} is computed by comparing the statistical distributions of the generated designs with those of real, professionally created designs from the training set. As shown in \cref{tab:table1}, the proposed method achieves a considerably lower FID score than the baseline model, indicating that it generates designs of higher quality and greater diversity, consistent with the findings of the qualitative study.

\subsection{Image-to-design generation}

\begin{figure*}[t]
  \centering
  \includegraphics[scale=0.55]{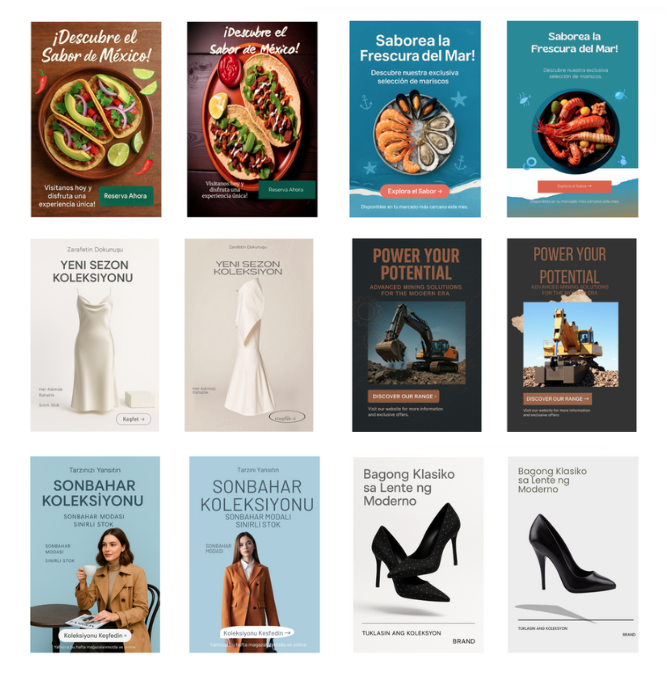}
  \caption{Qualitative results for the image-to-design task. In each pair, the left image shows the input generated by GPT-4o, and the right image shows the model’s output.}
  \label{fig:fig4}
\end{figure*}

In this experiment, the language model is replaced with a transformer with both encoder and decoder components, each consisting of 10 layers. The architecture has 500 million learnable parameters. Each design in the training set is first rendered, and the resulting image is divided into patches. These patch values are then concatenated with the CLIP embedding  \cite{b17} of the rendered image and fed into the transformer’s encoder. The model is trained to generate the corresponding design as a sequence of text interleaved with \texttt{[IMG]} tokens, following the approach used in the previous experiment. However, since replacing the language model with a plain transformer and using a relatively small training set makes learning natural language from scratch challenging, the content of each textbox is replaced with the word “TEXT” to simplify the task. Training is performed in float32 precision for 187,000 steps. Evaluation is conducted on a small set of graphic designs generated from curated prompts using GPT-4o \cite{b26}.

To evaluate the model's performance, both quantitative and qualitative studies are conducted. Cosine similarity is computed between the CLIP embedding of each original input image and the generated design to measure how closely the generated design matches the input. As shown in \cref{tab:table1}, the cosine similarity of CLIP embeddings is $0.85 \pm 0.05$, indicating strong adherence to the original images. \Cref{fig:fig4} illustrates qualitative examples produced by the trained model. In each pair, the left image is the GPT-4o generated input, and the right image is the model’s output. For easier assessment, the original text content is manually added to the generated text boxes. While minor inconsistencies in details are observed, the outputs generally align closely with the inputs. These results demonstrate that a single \texttt{[IMG]} token per image captures substantial semantic, stylistic, and color information. They also suggest that knowledge transferred from a pretrained language model has minimal effect on the model’s ability to map \texttt{[IMG]} tokens to CLIP embeddings. However, a pretrained language model is still required for generating the text content.

\section{Conclusion}

This paper introduces Giraffe, a novel architecture that maps hidden token representations from a language model to the embedding space of a visual model using a single \texttt{[IMG]} token per image. This approach allows language models to generate graphic designs, which would normally require very long sequences of interleaved text and image tokens. The mapping architecture consists of two shallow MLP blocks, each with a unique compression module followed by a shared expansion module that connects the blocks. The main block performs the mapping, while the second block assists in training to help the main block learn the mapping more effectively. During inference, the assisting block is removed, resulting in a lightweight solution. Experiments on a text-to-design task show that the architecture effectively captures the semantics, style, and color of each image, producing more complex designs with greater visual harmony than when images are represented solely through textual descriptions. The results also emphasize the importance of using a single token per image to achieve higher diversity and complexity in the generated graphic designs. Similarly, experiments on an image-to-design task demonstrate that the proposed architecture can capture semantics, style, and colour information with just one \texttt{[IMG]} token per image.

{
    \small
    \bibliographystyle{ieeenat_fullname}
    \bibliography{main}
}


\end{document}